\documentclass[runningheads]{llncs}
\usepackage{graphicx}
\usepackage{xcolor}
\begin{document}
\title{Observation Denoising in CYRUS Soccer Simulation 2D Team For RoboCup 2024}
\author{Nader Zare\inst{1} \and
Aref Sayareh\inst{2} \and
Sadra Khanjari \inst{3} \and
Arad Firouzkouhi \inst{4}
}
\authorrunning{N. Zare et al.}
\institute{CYRUS Robotic Team \and
Memorial University of Newfoundland, St. John's, Canada \and
Sharif University of Technology \and
Amirkabir University of Technology, Iran \\
\email{nader@cyrus2d.com}\\
\email{asayareh@mun.ca} \\
\email{arad.firouzkouhi@aut.ac.ir}\\ 
\url{https://cyrus2d.com/} \\
\url{https://github.com/Cyrus2D}
}
\maketitle              % typeset the header of the contribution
\begin{abstract}
In the Soccer Simulation 2D environment, accurate observation is crucial for effective decision-making. However, challenges such as partial observation and noisy data can hinder performance. To address these issues, we propose a denoising algorithm that leverages predictive modeling and intersection analysis to enhance the accuracy of observations. Our approach aims to mitigate the impact of noise and partial data, leading to improved gameplay performance. This paper presents the framework, implementation, and preliminary results of our algorithm, demonstrating its potential in refining observations in Soccer Simulation 2D. Cyrus 2D Team is using a combination of Helios, Gliders, and Cyrus base codes\cite{glbase,heliosbase,cyrusbase}.

\end{abstract}

\section{Introduction}
RoboCup has been organizing annual robotic soccer competitions since its inception in 1997, with the concept of robotic soccer games proposed as a novel research topic in 1992\cite{robo1997,noda1996soccer,kitano1997robocup}. These competitions aim to foster advancements in robotics and artificial intelligence by challenging participants to develop and implement robotic soccer games. To this end, various leagues have been established, including the rescue league, the soccer simulation league, and the standard platform soccer league.

The Soccer Simulation 2D (SS2D) league stands as one of the most established leagues in RoboCup. In this league, two teams, each comprising 11 players and a coach, compete in a virtual soccer game. Each match consists of 6000 cycles, equivalent to 10 minutes of real-time gameplay. During each cycle, players and coaches receive observations from the RoboCup Soccer Simulation Server and respond with appropriate actions. The server plays a crucial role in orchestrating the game, ensuring the smooth execution of each cycle.

The CYRUS team has been a consistent contender in RoboCup competitions since its debut in 2013. The team's crowning achievement came in 2021 when it won the championship title in the SS2D league at RoboCup. Alongside this victory, CYRUS has also secured second place in the years 2018, 2022, and 2023, and third place in 2019, demonstrating a strong and sustained presence in the competition. Additionally, the team has shown remarkable success in the Iran Open, clinching first place in 2014, 2018, 2021, and 2023, and taking second place in 2022. In the international arena, CYRUS emerged as the champion of the Asia-Pacific competition in 2018 and earned second place in the Japan Open in 2020.

In this paper, we will begin by reviewing related research conducted by other teams, with a focus on artificial intelligence and robotics in the context of soccer simulation 2D. Following this, we will highlight our previous works, showcasing our released applications and codes. We will then delve into the challenge of noise in soccer simulation 2D, presenting our innovative denoising algorithm designed to enhance observation accuracy. Finally, we will discuss our future plans for continued research and development in this field.

\section{Related Works}
This section highlights some of the recent innovative approaches developed by Soccer Simulation 2D (SS2D) teams:

The OXSY team enhanced their offensive strategy by focusing on increasing ball possession through accurate kick ball decisions, reacting proactively to events, moving effectively without the ball, creating more spaces behind the opponent's defense line, and receiving the ball in the most advantageous positions\cite{oxsy}.
Hermes2D team concentrated on improving their defense strategy by implementing a man-to-man marking algorithm, ensuring tighter coverage of opposing players\cite{hermes}.
The emperor team refined their Through Pass behavior using the Perceptron Neural Network Algorithm, optimizing the decision-making process for passing the ball\cite{emp}.
ITAndroids team focused on the optimization of field evaluator weights using the Covariance Matrix Adaptation Evolution Strategy (CMA-ES) to enhance team performance\cite{ita}.
RobôCIn team worked on a Dynamic Corner Marking strategy to bolster their defense, ensuring better coverage during corner situations\cite{robocin}.
The8 team improved their Pass Decision Making Method by employing a fuzzy algorithm based on the ball's position, enhancing the accuracy of passing decisions\cite{the8}.
Helios team's research centered around constructing tactical decisions for soccer simulation, with a particular focus on using learning-to-rank methods for ball-chasing behavior. Their approach aims to increase team flexibility and effectiveness in various tactical situations\cite{hel22,hel23}.

\subsection{Previous Works of CYRUS}
This section outlines the historical research and development journey of CYRUS since its establishment in 2013. The team has engaged in extensive research across various facets of the game, initially focusing on enhancing passing behavior, unmarking strategies, and team formations. Subsequently, the focus shifted towards refining defensive behavior, developing advanced shooting algorithms, and improving communication modules. A significant milestone was achieved in 2017 with the development of the opponent behavior prediction module. In recent years, the emphasis has been on advancing the pass prediction and unmarking module, observation de-noising by using LSTM, alongside the creation of the Data Generator framework\cite{cyrus13,cyrus14,cyrus15,cyrus16,cyrus17,cyrus18,cyrus19,cyrus21,cyrus22,cyrussamposiom,cyruschamp,cyrus22}.

\subsubsection{Cyrus 2014}
The CYRUS team made its source code from the 2014 RoboCup competition publicly available, providing a valuable resource for other teams and newcomers to the RoboCup community. This code offers insights into the team's strategies and implementations, aiding others in enhancing their performance. The source code is accessible on GitHub\footnote{Cyrus 2014 Source \url{https://github.com/naderzare/cyrus2014}.}.

\subsubsection{Cyrus2D Base}
The Cyrus2D Base code combines the best features of the Helios Base, Gliders Base, and the CYRUS team's winning strategies from the 2021 RoboCup Soccer Simulation 2D league. This base code serves as an ideal starting point for new teams embarking on soccer simulation projects, providing a solid foundation for developing necessary functionalities for the RoboCup competition. The code is available on GitHub\footnote{Cyrus Base Source \url{https://github.com/Cyrus2D/Cyrus2DBase}}.

\subsubsection{PYRUS Base}
The PYRUS Base code, developed in Python, offers a robust platform for researchers to develop and test Machine Learning algorithms within the soccer simulation context. This base code is particularly beneficial for teams looking to incorporate artificial intelligence techniques to improve performance. The code is available on GitHub\footnote{Pyrus Base \url{https://github.com/Cyrus2D/Pyrus-SS2D-Base}}.

\subsubsection{Cyrus AutoTest2D}
Initially developed by the WrightEagle team, the Cyrus AutoTest2D software has been enhanced and simplified by CYRUS team members. This software facilitates the running of parallel games to assess team performance, providing a systematic approach to evaluating strengths and weaknesses. The code is available on GitHub\footnote{AutoTest \url{https://github.com/Cyrus2D/AutoTest2D}}.

\subsubsection{AutoTune2D}
The AutoTune2D tool, developed by the CYRUS team, allows for fine-tuning of parameters in conjunction with AutoTest2D. This tool is invaluable for researchers aiming to optimize team performance by adjusting various parameters to find the optimal combination. The code is available on GitHub\footnote{AutoTune \url{https://github.com/Cyrus2D/AutoTune2D}}.

\subsubsection{Cyrus Log Analyzer}
The Cyrus Log Analyzer software is capable of analyzing Soccer Simulation 2D game logs to extract statistical values such as the number of passes and dribbles. This software provides crucial insights into team performance, aiding in the identification of areas for improvement and strategy refinement. The code is available on GitHub\footnote{CyrusLog Analyzer \url{https://github.com/Cyrus2D/CyrusLogAnalyzer}}.

\section{Observation Challenges in Soccer Simulation 2D}
In the realm of soccer simulation 2D, accurate observation plays a pivotal role in enabling players to make informed decisions on the field. However, the observation process consist of some challenges, primarily stemming from its partial and noisy nature.

\begin{itemize}
    \item \textbf{Partial Observation}: One of the foremost challenges lies in the partial nature of observation. In each cycle, players are limited in their field of vision, unable to perceive the entirety of the field at once. Consequently, players must strategically determine the optimal angle from which to survey their surroundings. This necessitates a nuanced understanding of the game dynamics, as players must continually adjust their viewing angles to glean critical information.
    \item \textbf{Noisy Observation}: 
    Compounding the challenge is the inherent noise present in the observation process. The information provided to players regarding the distance and direction of objects is subject to quantization, introducing inaccuracies into their perceptions of the field. Specifically, angles are rounded to the nearest integer degree, while distances are quantized based on the relative positions of the object and the observer. Notably, as the distance between players and objects increases, the level of noise escalates, further exacerbating the difficulty of accurate observation.
\end{itemize}

The ramifications of these observation challenges are profound, profoundly impacting gameplay dynamics. Accurate knowledge of the ball's position empowers players to execute precise kicks and interceptions, enhancing overall gameplay fluidity. Similarly, precise awareness of players' positions facilitates more informed decision-making, reducing the inherent risks associated with strategic maneuvers such as passing, shooting, and dribbling. Thus, the quest for enhanced observation accuracy stands as a critical endeavour in optimizing performance and elevating the quality of soccer simulation 2D.

\section{Denoising Observation in Soccer Simulation 2D}

In the pursuit of enhancing observation accuracy amidst the challenges of partial and noisy data, our approach incorporates a denoising algorithm designed to refine the perception of players and the ball's positions on the field. Leveraging predictive modelling and intersection analysis, our denoising strategy aims to extrapolate the true positions of objects based on observed data and movement dynamics.

% Predictive Modeling:
Central to our denoising methodology is the utilization of predictive modelling to anticipate the movement of objects, namely players and the ball. When a player receives observation data indicating the presence of an object within a particular sector, we initiate a predictive process. For the ball, we incorporate its velocity into our predictions, incrementally extending the predicted area with each cycle to account for potential movement. Despite inherent inaccuracies in velocity angle and length, this predictive approach enables us to forecast the probable trajectory of the ball.

% Intersection Analysis:
At each subsequent cycle, we compare the newly observed sector with the predicted area generated in the previous cycle. By identifying the intersection of these regions, we refine our estimation of the object's position. This intersection analysis serves as a denoising mechanism, mitigating the effects of noise inherent in the observation process. The midpoint of the intersection serves as our updated estimate of the object's position, offering a more precise representation amidst the partial and noisy observation data.

% Player-Specific Prediction:
In addition to ball movement prediction, our denoising algorithm incorporates player-specific prediction mechanisms. At the onset of the game, we compile a table for each player type, detailing the maximum pass distance achievable with optimal power across various angles. Leveraging this precomputed information, we extrapolate the potential movement of players during cycles where direct observation is unavailable. By referencing the relevant player type table, we estimate the probable positions of players and refine these estimates through intersection analysis with observed sectors in subsequent cycles.

% Iterative Refinement:
Given the partial nature of observation, our denoising algorithm operates iteratively, continuously refining object positions based on updated observation data. In instances where direct observation is not feasible, we extend predicted areas and iteratively intersect them with newly observed sectors, progressively enhancing the accuracy of our estimations over time.

\begin{figure}
    \centering
    \includegraphics[width=1\textwidth]{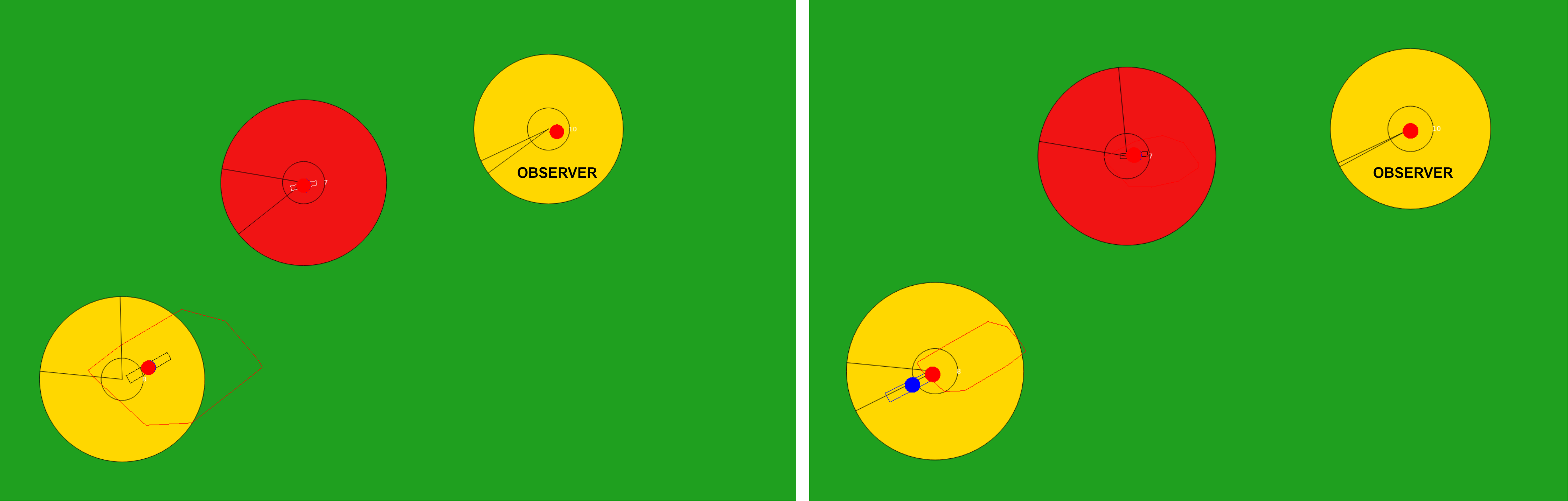}
    \caption{The observer is denoising its teammate's position based on our denoising model. The left image is the previous cycle of the right image. the red polygon shows the prediction of our model, while the black sector shows the information that we receive from the server. The blue dot shows the position of the teammate without any denoising, and the red dot shows the position of the player with our new model. In the left image, the blue dot is hidden under the red dot.}
    \label{fig:enter-label}
\end{figure}

As Fig \ref{fig:enter-label} shows, the is looking at its teammate for two consecutive cycles. In the first cycle, our denoising model predicted an area around the base prediction, which the intersection was as same as the base denoising model. However, at the next cycle, the teammate moved while our observer received new information about its position. Our denoising model now predicted an area with a lower common area with the base denoising model. Therefore, the intersection area had a lower area which enabled the observer to accurately find the new position of the player.

In our experimental evaluations, the implementation of the denoising observation algorithm yielded promising results, demonstrating notable improvements in the accuracy of player coordination determination within the soccer simulation 2D environment. On average, our agents achieved a substantial enhancement in pinpointing the positions of opposing players, surpassing the baseline coordination by an average of $5.37cm$. This significant improvement underscores the efficacy of our denoising approach in mitigating the adverse effects of observation noise and partial visibility, thereby facilitating more precise decision-making and strategic gameplay execution.

\section{Conclusion and Future Works}

In conclusion, our investigation into denoising observation techniques within soccer simulation 2D has yielded promising outcomes, showcasing the potential for substantial enhancements in player coordination accuracy. By leveraging predictive modeling and intersection analysis, our algorithm effectively mitigates the detrimental effects of partial visibility and observation noise, enabling agents to make more informed decisions on the field. The demonstrated improvement of 5.37 centimeters in player coordination accuracy signifies a significant advancement in the realm of soccer simulation, underscoring the practical applicability and efficacy of our denoising approach.

Looking ahead, there are several avenues for further exploration and refinement of denoising observation techniques in soccer simulation 2D. Firstly, enhancing the predictive capabilities of our algorithm, particularly in accounting for dynamic changes in player and ball movement, could yield even greater improvements in coordination accuracy. Additionally, exploring advanced machine learning methodologies, such as deep learning-based approaches, may offer novel insights and optimizations for denoising observation in dynamic and complex environments. Furthermore, extending our research to incorporate real-time adaptation and learning mechanisms could enable agents to adapt more effectively to evolving game scenarios, thereby further enhancing their performance and competitiveness. Overall, continued research and development in this domain hold the potential to revolutionize the landscape of soccer simulation, paving the way for more immersive and engaging gameplay experiences.


\begin{thebibliography}{8}
\bibitem{glbase}
Prokopenko, M., Wang, P.: Gliders2d: Source Code Base for RoboCup 2D Soccer Simulation League. CoRR abs/1812.10202 (2018)

\bibitem{heliosbase}
Akiyama, H., Nakashima, T.: Helios base: An open source package for the robocup soccer 2d simulation. In Robot Soccer World Cup 2013 Jun 24 (pp. 528-535). Springer, Berlin, Heidelberg.

\bibitem{cyrusbase}
Zare, N., Amini, O., Sayareh, A., Sarvmaili, M., Firouzkouhi, A., Ramezani Rad, S., Matwin, S., Soares, A.: Cyrus2D base: Source Code Base for RoboCup 2D Soccer Simulation League. In: RoboCup 2022: Robot World Cup XXV, Springer (2022)

\bibitem{robo1997}
Kitano, H., Asada, M., Kuniyoshi, Y., Noda, I. and Osawa, E., 1997, February. Robocup: The robot world cup initiative. In Proceedings of the first international conference on Autonomous agents (pp. 340-347).

\bibitem{noda1996soccer}
Noda, I. and Matsubara, H., 1996, November. Soccer server and researches on multi-agent systems. In Proceedings of the IROS-96 Workshop on RoboCup (pp. 1-7).

\bibitem{kitano1997robocup}
Kitano, H., Asada, M., Kuniyoshi, Y., Noda, I., Osawa, E. and Matsubara, H., 1997. RoboCup: A challenge problem for AI. AI magazine, 18(1), pp.73-73.

\bibitem{oxsy}
Marian, S., Luca, D., Sacuiu, R., Sarac, B., Cotarlea, O.: OXSY 2023 Team Description. In: RoboCup 2023 Symposium and Competitions. France (2023).

\bibitem{hermes}
Sana Abedi, Tara Aghaei, Fatemeh Ghasemi, Parinaz Rastegar, Fatemeh Maleki, Parima Tghados, Nazanin Sarlak, Rozhina Pourmoghaddam: Hermes2D Soccer2D simulation Team Description Paper. In: RoboCup 2023 Symposium and Competitions. France (2023).

\bibitem{emp}
Fathi, E., Mazloum, S.: EMPEROR Soccer Simulation 2D Team Description Paper 2023. In: RoboCup 2023 Symposium and Competitions. France (2023).

\bibitem{ita}
Davi M. Vasconcelos, Nean Segura, and Marcos R. O. A. Maximo: ITAndroids 2D Soccer Simulation Team Description Paper 2023. In: RoboCup 2023 Symposium and Competitions. France (2023).

\bibitem{robocin}
Cristiano Santos de Oliveira1, Mateus Gon calves Machado, et al: RoboCIn Team Description Paper 2023. In: RoboCup 2023 Symposium and Competitions. France (2023).

\bibitem{the8}
Noohpisheh, M., Shekarriz, M., Nematollahi, R., Ghasemi, F., Mohammadi, M., Amiri, N., Amiri, S.: The8 Soccer 2D Simulation Team Description Paper 2023. In: RoboCup 2023 Symposium and Competitions. France (2023).

\bibitem{hel22}
Akiyama, H., Nakashima, T., Hatakeyama, K.: HELIOS2022: Team Description Paper. In: RoboCup 2022 Symposium and Competitions. Thailand (2022).

\bibitem{hel23}
Akiyama, H., Nakashima, T., Hatakeyama, K., Fujikawa, T.: HELIOS2023: Team Description Paper. In: RoboCup 2023 Symposium and Competitions. France (2023).


\bibitem{cyrus13}
Khayami, R., Zare, N., Zolanvar, H. M. Karimi, P., Mahor, F., Tekara, E., Asali, Fatehi, M. : Cyrus Soccer 2D Simulation Team Description Paper 2013. In The 17th annual RoboCup International Symposium, Eindhovenm, The Netherlands. (2013)

\bibitem{cyrus14}
Khayami, R., Zare, N., Karimi, M., Mahor, P., Afshar, A., Najafi, M. S., Asadi, M., Tekrar, F., Asali, E., Keshavarzi, A.: CYRUS 2D simulation team description paper 2014. In: RoboCup 2014. Joao Pessoa, Brazil, (2014).

\bibitem{cyrus15}
Zare, N., Karimi, M., Keshavarzi, A., Asali, E., Ali Poor, H., Aminian, A., Beheshtian, E., Mola, H., Jafari, H. , Khademian, M. J.: Cyrus Soccer 2D Simulation Team Description Paper 2015. In: RoboCup 2015. Hefei, China, (2015).

\bibitem{cyrus16}
Zare, N., Keshavarzi, A., Beheshtian, S. E., Mowla, H., Akbarpour, A., Jafari, H., Arab Baraghi, K., Zarifi, M. A., Javidan, R.: Cyrus 2D Simulation Team Description Paper 2016. In: RoboCup 2016. Leipzig, Germany, (2016).

\bibitem{cyrus17}
Zare, N., Najimi, A., Sarvmaili, M., Akbarpour, A., NaghipourFar, M., Barahimi, B., Nikanjam, A.: Cyrus 2D Simulation Team Description Paper 2017, In: Robocup(2017), Hefei, China (2017).

\bibitem{cyrus18}
Zare, N., Sadeghipour, M., Keshavarzi, A., Sarvmaili, M., Nikanjam, A., Aghayari, R., Firouzkoohi, A., Abolnejad, M., Elahimanesh, S., Akhgari, A.: Cyrus 2D Simulation Team Description Paper 2018. In: RoboCup(2018), Montreal, Canada (2018).

\bibitem{cyrus19}
Zare, N., Sarvmaili, M., Mehrabian, O., Nikanjam, A., Khasteh, S.-H., Sayareh, A., Amini, O., Barahimi, B., Majidi, A., Mostajeran, A.: Cyrus 2D Simulation 2019. In: RoboCup (2019).

\bibitem{cyrus21}
Zare, N., Sayareh, A., Sarvmaili, M., Amini, O., Soares, A., Matwin, S.: CYRUS 2D Soccer Simulation Team Description Paper 2021. In: RoboCup 2021 Symposium and Competitions, Worldwide (2021)

\bibitem{cyrus22}
Zare, N., Firouzkouhi, A., Amini, O., Sarvmaili, M., Sayareh, A., Soares, A., Matwin, S.: CYRUS 2D Soccer Simulation Team Description Paper 2022. In: RoboCup 2022 Symposium and Competitions, Thailand (2022)

\bibitem{cyrussamposiom}
Zare, N., Sayareh, A., Sarvmaili, M., Amini, O., Matwin, S., Soares, A.: Engineering Features to Improve Pass Prediction in 2D Soccer Simulation Games. In: RoboCup 2021: Robot World Cup XXIV, Springer (2021)

\bibitem{cyruschamp}
Zare, N., Amini, O., Sayareh, A., Sarvmaili, M., Firouzkouhi, A., Matwin, S., Soares, A.: Improving Dribbling, Passing, and Marking Actions in Soccer Simulation 2D Games Using Machine Learning. In: RoboCup 2021: Robot World Cup XXIV, Springer (2021)



\bibitem{cyrus23}
Sayareh, A., Zare, N., Amini, O., Firouzkouhi, A., Sarvmaili, M. and Matwin, S., 2023. Observation Denoising in CYRUS Soccer Simulation 2D Team For RoboCup 2023. In: RoboCup 2023 Symposium and Competitions, France (2023). arXiv preprint arXiv:2305.19283.


\end{thebibliography}
%

\end{document}